\newcommand{\specialcell}[2][l]{%
  \begin{tabular}[#1]{@{}l@{}}#2\end{tabular}}
  
\documentclass[11pt,a4paper]{article}

\usepackage[hyperref]{naaclhlt2018}
\usepackage{times}
\usepackage{latexsym}
\usepackage{float}
\usepackage{graphicx}
\usepackage{subfigure}
\usepackage{comment}
\usepackage{url}
\aclfinalcopy 
\def\aclpaperid{546} 
\usepackage{caption}

\DeclareCaptionFont{10pt}{\fontsize{10pt}{12pt}\selectfont}
\title{Exploring the Role of Prior Beliefs for Argument Persuasion}
\author{Esin Durmus \\
 Cornell University \\ 
{\tt ed459@cornell.edu} \And Claire Cardie \\ Cornell University \\ {\tt cardie@cs.cornell.edu}}
\date{}
\begin{document}
\maketitle
\begin{abstract}
  Public debate forums provide a common platform for exchanging opinions on a topic of interest. While recent studies in  natural language processing (NLP) have provided empirical evidence that the language of the debaters and their patterns of interaction play a key role in changing the mind of a reader, research in psychology has shown that prior beliefs can affect our interpretation of an argument and could therefore constitute a competing alternative explanation for resistance to changing one's stance. To study the actual effect of language use vs.\ prior beliefs on persuasion, we provide a new dataset and propose a controlled setting that takes into consideration two reader-level factors: political and religious ideology. We find that prior beliefs affected by these reader-level factors play a more important role than language use effects and argue that it is important to account for them in NLP studies of persuasion.
\end{abstract}

\section{Introduction}

Public debate forums provide to participants a common platform for expressing their 
point of view on a topic;  they also present to participants the different sides of an argument. 
The latter can be particularly important:
awareness of divergent points of view allows one, in theory, to make a fair and informed decision
about an issue; and exposure to new points of view can furthermore possibly persuade a reader to change 
his overall stance on a topic. 

Research in natural language processing (NLP) has begun to study persuasive
writing and the role of language in persuasion. \newcite{tan2016winning} and \newcite{zhang2016conversational},
for example, have shown that the language of
opinion holders or debaters and their patterns of interaction play a key role in changing the mind of a reader. 
At the same time, research in psychology has shown that prior beliefs can
affect our interpretation of an argument
even when the argument consists of numbers and empirical 
studies that would seemingly belie misinterpretation \cite{lord1979biased,vallone1985hostile,doi:10.1177/0741088396013003001}.

We hypothesize that studying the actual effect of language on
persuasion will require a more controlled experimental setting --- one that takes into account
any potentially confounding user-level (i.e., reader-level) factors\footnote{Variables that affect both the dependent and 
independent variables causing misleading associations.} 
that could cause a person to change, or keep a person from changing, his opinion.
In this paper we study one such type of factor: the prior beliefs
of the reader as impacted by their political or religious ideology.
We adopt this focus 
since it has been shown that 
ideologies play an important role for an individual when they form beliefs 
about controversial topics, and potentially affect how open the individual 
is to being persuaded \citep{stout1996religion, doi:10.1111/j.1540-5907.2005.00161.x, croucher2012religion}.

We first present a dataset of online debates that enables us to construct
the setting described above in which we can study the effect of language on persuasion while taking
into account selected user-level factors.
In addition to the text of the debates, the dataset contains a multitude of
background information on the users of the debate platform.
To the best of our knowledge, it is the first publicly available dataset of debates
that simultaneously provides such comprehensive information about the debates, the debaters and those 
voting on the debates.

With the dataset in hand, we then propose the novel task of studying persuasion 
(1) at the level of individual users, and (2) in a setting that can control for 
selected user-level factors, in our case, the prior beliefs associated with the 
political or religious ideology of the debaters and voters.
In particular, previous studies focus on predicting the winner of a debate based on the cumulative change 
in pre-debate vs.\ post-debate votes for the opposing sides \cite{zhang2016conversational,potash2017towards}.
In contrast, we aim to 
predict which debater an individual user (i.e., reader of the debate) perceives as more successful, given their stated political and religious ideology.

Finally, we identify which features appear to be most important for persuasion,
considering the selected user-level factors as well as the more traditional linguistic 
features associated with the language of the debate itself. We hypothesize that 
the effect of political and religious ideology will be stronger when the debate topic 
is \textit{Politics} and \textit{Religion}, respectively.
To test this hypothesis, we experiment with debates on only \textit{Politics}
or only \textit{Religion} vs.\ debates from all topics including \textit{Music}, 
\textit{Health}, \textit{Arts}, etc.
 
Our main finding is that prior beliefs associated with the selected user-level factors
play a larger role than linguistic features when predicting the successful debater in a debate. In addition, the effect of these factors varies 
according to the topic of the debate topic.
The best performance, however, is achieved when we 
rely on features extracted from user-level factors in conjunction with linguistic features 
derived from the debate text.
Finally, we find that the set of linguistic features that emerges as the most predictive
changes when we control for user-level factors (political and religious ideology)
vs.\ when we do not, showing the importance of accounting for these factors 
when studying the effect of language on persuasion.

In the remainder of the paper, we describe the debate dataset (Section~\ref{data})
and the prediction task (Section~\ref{task}) followed by the experimental
results and analysis (Section~\ref{results}), related work (Section~\ref{rel-work})
and conclusions (Section~\ref{conclusions}).

\section{Dataset}
\label{data}
For this study, we collected $67,315$ debates 
from debate.org\footnote{\href{http://www.debate.org}{www.debate.org}}
from $23$ different topic
categories including \textit{Politics}, \textit{Religion}, \textit{Health}, \textit{Science} and \textit{Music}\footnote{The
dataset will be made publicly available at \href{http://www.cs.cornell.edu/~esindurmus/}{http://www.cs.cornell.edu/~esindurmus/}.}. In addition
to text of the debates, we collected $198,759$ votes from the readers of these debates. Votes evaluate different dimensions of the debate. 

To study the effect of user characteristics, we collected user information for
$36,294$ different users. Aspects of the dataset most relevant to our task are explained in the following section in more detail. 
\subsection{Debates}

\begin{figure}
\centering
\includegraphics[width=7.8cm,height=4.1cm]{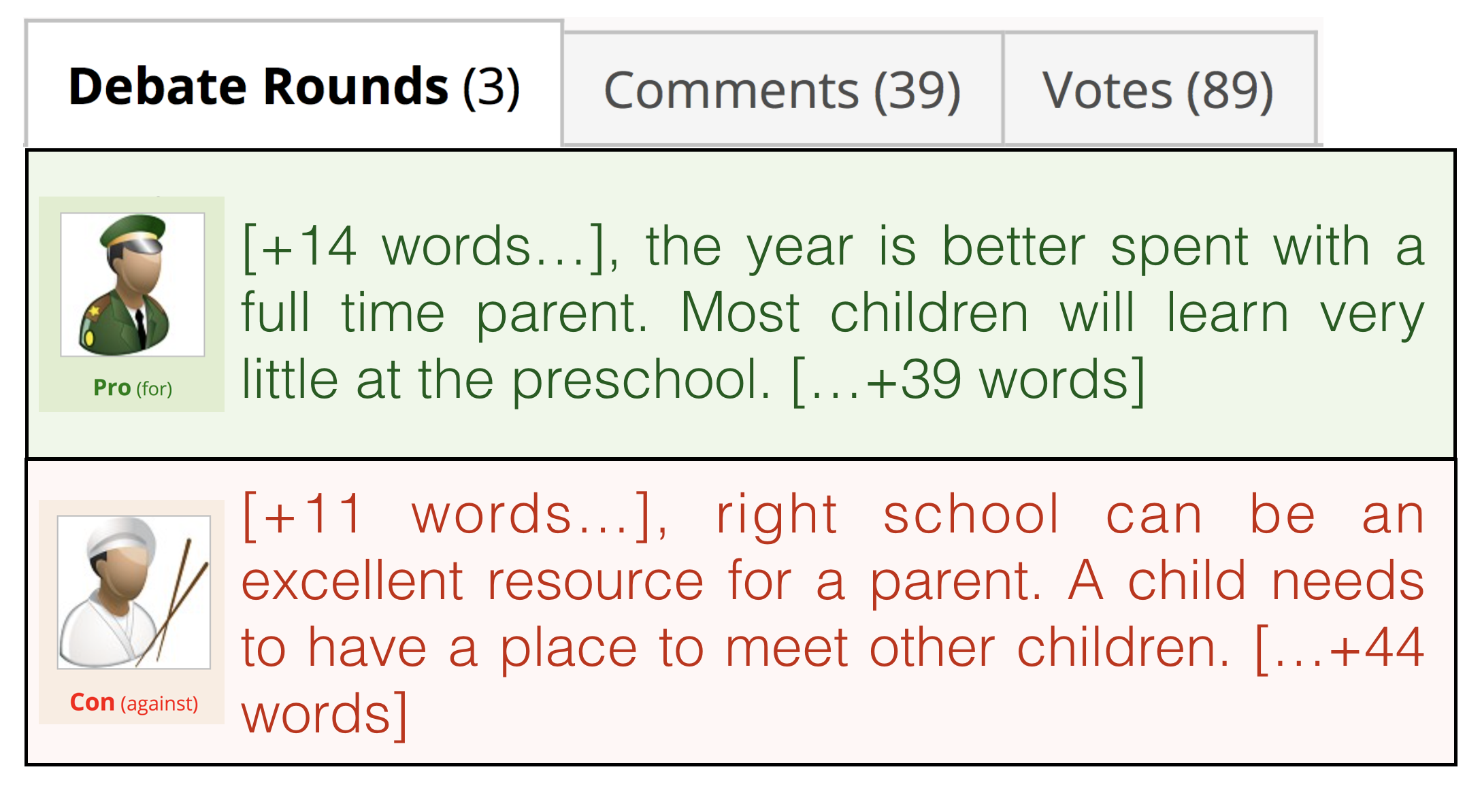}
\caption{{\sc round 1} for the debate claim \sc {``Preschool Is A Waste Of Time".}}
\label{figure_round}
\end{figure}

\hspace{1em}\textbf{Debate rounds.}
Each debate consists of a sequence of {\sc rounds} in which two debaters from opposing sides (one is supportive of the claim (i.e., {\sc pro}) and the other is against the claim (i.e., {\sc con})) provide their arguments. Each debater has a single chance in a {\sc round} to make his points. Figure \ref{figure_round} shows an example {\sc round 1}  for the debate claim {\sc ``Preschool Is A Waste Of Time"}. 
The number of {\sc rounds} in debates ranges from $1$ to $5$ and the majority of debates ($61,474$ out of $67,315$) contain $3$ or more {\sc rounds}.
\noindent

\begin{figure}
\includegraphics[width=8cm,height=4.2cm]{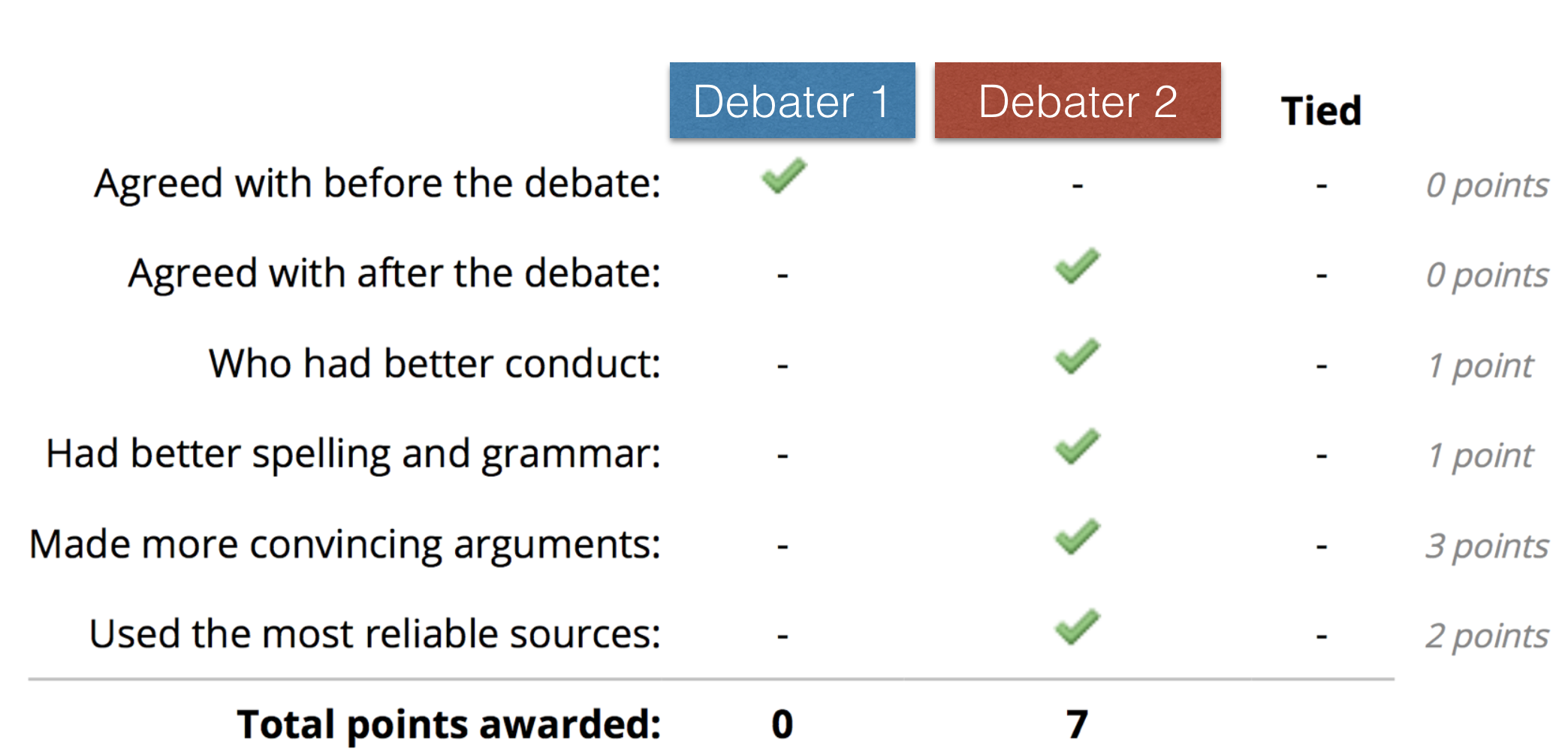}
\caption{An example post-debate vote. Convincingness of arguments contributes to the total points the most.}
\label{figure_vote}
\end{figure}

\textbf{Votes.}
All users in the \textit{debate.org} community can vote on debates. As shown in Figure \ref{figure_vote}, voters share their stances on the debate topic before and after the debate and evaluate the debaters' conduct, their spelling and grammar, the convincingness of their arguments and the reliability of the sources they refer to. For each such dimension, voters have the option to choose one of the debaters as better or indicate a tie. 
This fine-grained voting system gives a glimpse into the reasoning behind the voters' decisions. \\

\subsubsection{Determining the successful debater}
There are two alternate criteria for determining the successful debater in a debate. 
Our experiments consider both. 

\textbf{Criterion 1: Argument quality.}
As shown in Figure 2, debaters get points for each dimension of the debate. The most important dimension --- in that it contributes most to the point total --- is making convincing arguments. \textit{debate.org} uses Criterion 1 to determine the winner of a debate. 

\textbf{Criterion 2: Convinced voters.} Since voters share their stances before and after the debate, the debater who convinces more voters to change their stance is declared as the winner.

\subsection{User information} 
\begin{figure}
\includegraphics[width=6.6cm,height=5.6cm]{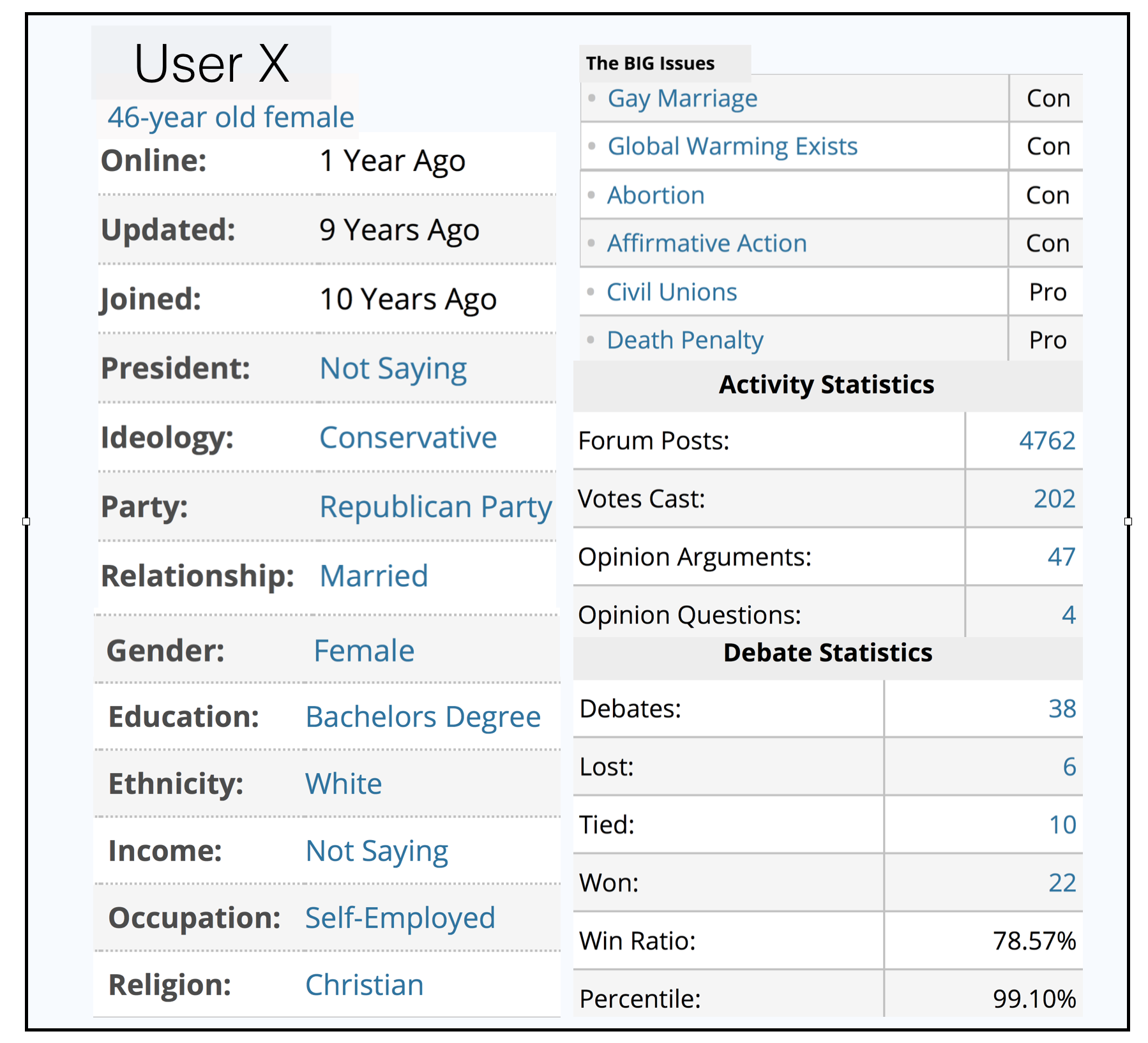}
\caption{An example of a (partial) user profile.\\
Top right: Some of the  \textit{big issues} on which the user shares his opinion are included. The user is against ({\sc con}) abortion and gay marriage and in favor of ({\sc pro}) the death penalty.}
\label{figure_user}
\end{figure}

On \textit{debate.org}, each user has the option to share demographic and private state information such as their age, gender, ethnicity, political ideology, religious ideology, income level, education level, the president and the political party they support. Beyond that, we have access to information about their activities on the website such as their overall success rate of winning debates, the debates they participated in as a debater or voter, and their votes. An example of a user profile is shown in Figure \ref{figure_user}.

\textbf{Opinions on the \textit{big issues}.}
\textit{debate.org} maintains a list of the most controversial debate topics as determined  by the editors of the website. These are referred to as \textit{big issues}.\footnote{\label{big_issues}\href{http://www.debate.org/big-issues/}{http://www.debate.org/big-issues/}} 
Each user shares his stance on each \textit{big issue} on his profile (see Figure \ref{figure_user}): either {\sc pro} (in favor), {\sc con} (against), {\sc n/o} (no opinion), {\sc n/s} (not saying) or {\sc und} (undecided).

\begin{figure}[h]
\includegraphics[width=6cm,height=5cm]{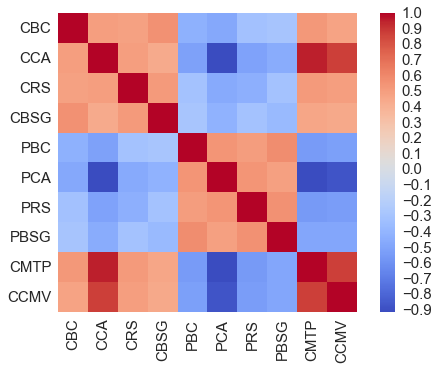}
\caption{The correlations among argument quality dimensions.}
\label{voting_dimension}
\end{figure}

\section{Prediction task: which debater will be  declared as more successful by an individual voter?}
\label{task}
In this section, we first analyze which dimensions of argument quality are the most important for determining the successful debater. Then, we analyze whether there is any connection between selected user-level factors and users' opinions on the \textit{big issues} to see if we can infer their opinions from these factors. Finally, using our findings from these analyses, we perform the task of predicting which debater will be perceived as more successful by an individual voter. 

\subsection{Relationships between argument quality dimensions}

Figure \ref{voting_dimension} shows the correlation between pairs of voting dimensions (in the
first 8 rows and columns) and the correlation of each dimension with (1) getting more points 
(row or column 9) and (2) convincing more people as a debater (final row or column).
Abbreviations stand for (on the {\sc con} side): has better conduct ({\sc cbc}), 
makes more convincing arguments ({\sc cca}), uses more 
reliable sources ({\sc crs}), has better spelling and grammar ({\sc cbsg}), gets more total points ({\sc cmtp}) and
convinces more voters ({\sc ccmv}). For the {\sc pro} side we use {\sc pbc}, {\sc pca}, and so on. 

From Figure \ref{voting_dimension}, we can see that making more convincing arguments ({\sc cca}) correlates the most with total points ({\sc cmtp}) and
convincing more voters ({\sc ccmv}). 
This analysis motivates us to identify the linguistic features that are indicators of more 
convincing arguments.

\subsection{The relationship between a user's opinions on the \textit{big issues} and their prior beliefs}
We disentangle different aspects of a person's prior beliefs to understand how well each correlates with their opinions on the \textit{big issues}. 
As noted earlier, we focus here only on prior beliefs in the form of self-identified political and religious ideology.

\begin{figure}
\includegraphics[width=7.8cm,height=2cm]{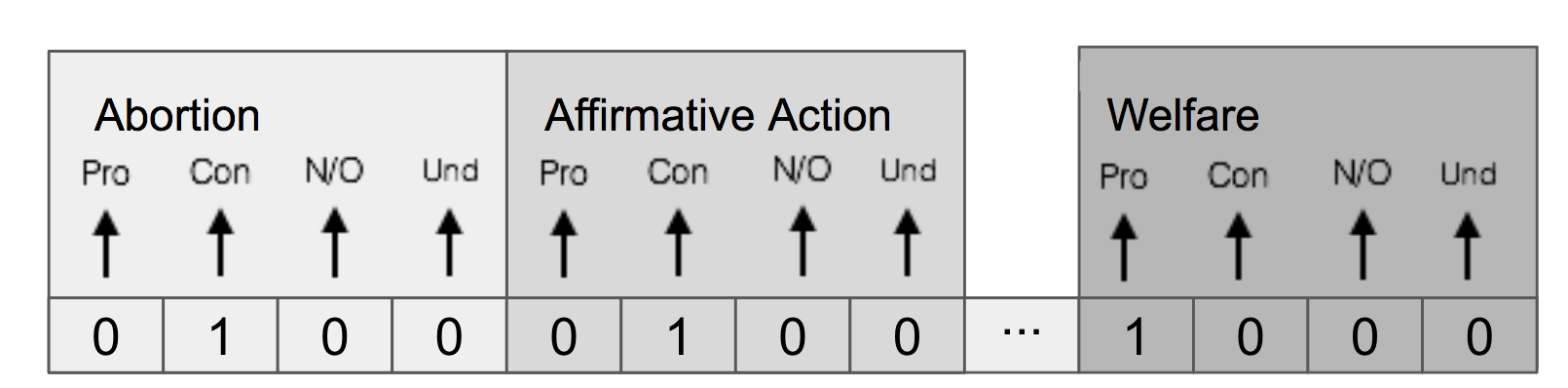}
\caption{The representation of the {\sc BigIssues} vector derived by this user's decisions on \textit{big issues}. Here,  the user is {\sc con} for {\sc abortion} and {\sc affirmative action} issues and {\sc pro} for the {\sc welfare} issue.}
\label{figure_big_issues}
\end{figure}

\textbf{Representing the \textit{big issues}.} To represent the opinions of a user on a \textit{big issue}, we use a four-dimensional one-hot encoding where the indices of the vector correspond to {\sc pro}, {\sc con}, {\sc n/o} (no opinion), and {\sc und} (undecided), consecutively (1 if the user chooses that value for the issue, 0 otherwise). Note that we do not have a representation for {\sc n/s} since we eliminate users having {\sc n/s} for at least one \textit{big issue} for this study.  We then concatenate the vector for each \textit{big issue} to get a representation for a user's stance on all the \textit{big issues} as shown in Figure \ref{figure_big_issues}. We denote this vector by {\sc BigIssues}.
%

We test the correlation between the individual's opinions on \textit{big issues} and the selected user-level factors in this study using two different approaches: clustering and classification.
\begin{figure}
  \centering
  \subfigure[{\sc liberal} vs. {\sc conservative}
  ]{\includegraphics[scale=0.41]{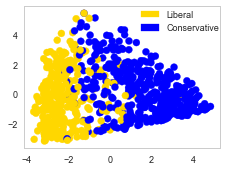}.}\quad
  \subfigure[{\sc atheist} vs. {\sc christian}.]{\includegraphics[scale=0.41]{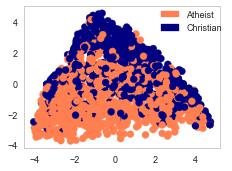}}
  \caption{PCA representation of decisions on \textit{big issues} color-coded with political and religious ideology. We see more distinctive clusters for {\sc conservative} vs. {\sc liberal} users suggesting that people's opinions are more correlated with their political ideology.}
\label{clustering_figure}
\end{figure}

\textbf{Clustering the users' decisions on \textit{big issues}.} We apply PCA on the {\sc BigIssues} vectors
 of users who identified themselves as {\sc conservative} vs.\ {\sc liberal} ($740$ users). We do the same for the users who identified
 themselves as {\sc atheist} vs.\ {\sc christian} ($1501$ users).
 In Figure \ref{clustering_figure}, we see that there are distinctive clusters of {\sc conservative} vs.\ {\sc liberal} users in the two-dimensional representation while for {\sc atheist} vs.\ {\sc christian}, the separation is not as distinct. This suggests that people's opinions on the \textit{big issues} identified by \textit{debate.org} correlate more with their political ideology than their religious ideology.

\textbf{Classification approach.} We also treat this as a classification task\footnote{For all the classification tasks described in this paper, we experiment with logistic regression, optimizing the regularizer ($\ell$1 or $\ell$2) and the
regularization parameter C (between $10^{-5}$ and $10^{5}$).} using the {\sc BigIssues} vectors for each user as 
features and the user's religious and political ideology as the labels to be predicted.
So the classification task is: Given the user's {\sc BigIssues} vector, predict his political and religious ideology. Table \ref{belief_table}
shows the accuracy for each case. We see that using the {\sc BigIssues} vectors as features performs
significantly better\footnote{\label{significance}We performed the McNemar significance test.} than majority 
baseline\footnote{The majority class baseline predicts {\sc conservative} for political and {\sc christian} for religious ideology for each example, respectively.}.

\begin{table}
\begin{center}
\begin{tabular}{|l|rl|}
\hline \bf Prior belief type & \bf Majority & \bf \sc{BigIssues} \\ \hline
Political Ideology & $57.70$\% & $92.43$\% \\
Religious Ideology & $52.70$\% & $82.81$\% \\
\hline
\end{tabular}
\end{center}
\caption{\label{font-table} Accuracy using majority baseline vs.\ {\sc BigIssues} vectors as features.}
\label{belief_table}
\end{table}

This analysis shows that there is a clear relationship between people's opinions on the \textit{big issues} and the selected user-level factors. 
It raises the question of whether it is even possible to persuade someone with prior beliefs relevant
to a debate claim to change their stance on the issue. It may be the case that people prefer to agree with the individuals having the same (or similar) beliefs regardless of the quality of the arguments and the particular language used.
Therefore, it is important to understand the relative effect of prior beliefs
vs.\ argument strength on persuasion. 

\subsection{Task descriptions}
Some of the previous work in NLP on persuasion focuses on predicting the winner of a debate as determined by the change in the number of people supporting each stance before and after the debate \cite{zhang2016conversational,potash2017towards}. However, we believe that studies of the effect of language on persuasion should take into account other, extra-linguistic, factors that can affect opinion change: in particular, we propose an experimental framework for studying the effect of language on persuasion that aims to control for the prior beliefs of the reader as denoted through their self-identified political and religious ideologies.
As a result, we study a more fine-grained prediction task: for an individual voter, predict which 
side/debater/argument the voter will declare as the winner.

\begin{table*}[h]
\begin{center}
\begin{tabular}{p{5cm}p{8cm}}
\hline
\textbf{User-based features}&\textbf{Description} \\
    \hline
    \textbf{Opinion similarity.} & For $userA$ and $userB$, the cosine similarity of {\sc BigIssues$_{userA}$} and {\sc BigIssues$_{userB}$}.\\
    \textbf{Matching features.} & For $userA$ and $userB$, 1 if $userA_f$==$userB_f$, 0 otherwise where
$f$ $\in$ $\{$political ideology, religious ideology$\}$. We denote these features as \textit{matching political ideology} and \textit{matching religious ideology}.\\
    \hline
    \textbf{Linguistic features} & \textbf{Description} \\
    \hline
    \textbf{Length.} &  Number of tokens.\\
    \textbf{Tf-idf.} & Unigram, bigram and trigram features. \\
    \textbf{Referring to the opponent.} &  Whether the debater refers to their opponent using words or phrases like ``opponent, my opponent''.\\
   \textbf{Politeness cues.} &  Whether the text includes any signs of politeness such as ``thank'' and ``welcome''.\\
    \textbf{Showing evidence.} & Whether the text has any signs of citing any other sources (e.g., phrases like ``according to''), or quotation.\\
    \textbf{Sentiment.} &  Average sentiment polarity. \\
    \specialcell[b]{\textbf{Subjectivity}  \cite{wilson2005recognizing}.} &  Number of words with negative strong, negative weak, positive strong, and positive weak subjectivity.\\ 

    \textbf{Swear words.} & \# of swear words. \\
    \textbf{Connotation score}
    \cite{feng2011classifying}. &  Average \# of words with positive, negative and neutral connotation.\\
    \textbf{Personal pronouns.} & Usage of first, second, and third person pronouns. \\
    \textbf{Modal verbs.} &  Usage of modal verbs.\\
    \textbf{Argument lexicon features.} \cite{somasundaran2007detecting}. & \# of phrases corresponding to different argumentation styles. \\
    \textbf{Spelling.} &  \# of spelling errors.\\
    \textbf{Links.} & \# of links. \\
    \textbf{Numbers.} &  \# of numbers. \\
    \textbf{Exclamation marks.} &  \# of exclamation marks. \\
    \textbf{Questions.} &  \# of questions. \\
   
  \hline
\end{tabular}
\end{center}
\caption{Feature descriptions}
\label{table:features}
\end{table*}

\textbf{Task 1 : Controlling for religious ideology.}
In the first task, we control for religious ideology by selecting debates for which each of the two
debaters is from a different religious ideology (e.g., debater 1 is {\sc atheist}, debater 2 is 
{\sc christian}).  In addition, we consider only voters that (a) self-identify with one of these 
religious ideologies (e.g., the voter is either {\sc atheist} or {\sc christian}) and (b) 
changed their stance on the debate claim post-debate vs.\ pre-debate.  
For each such voter, we want to predict which of the {\sc pro}-side debater
or the {\sc con}-side debater did the convincing.  Thus, in this task, we use 
\textit{Criterion 2} to determine the winner of the debate from the point of
view of the voter.  Our hypothesis is that the voter will be convinced by the
debater that espouses the religious ideology of the voter.

In this setting, we can study the factors that are important for a particular voter to be convinced by a debater. This setting also provides an opportunity to understand how the voters who change their minds perceive arguments from a debater who is expressing the same 
vs.\ the opposing prior belief.

To study the effect of the debate topic, we perform this study for two cases --- debates belonging to the \textit{Religion} category and then all the categories.  The \textit{Religion} category contains debates like {\sc ``Is the Bible against women's rights?'' } and {\sc ``Religious theories should not be taught in school''}. We want to see how strongly a user's religious ideology affects the persuasive effect of language in such a topic as compared to the all topics.  We expect to see stronger effects of prior
beliefs for debates on \textit{Religion}.

\textbf{Task 2: Controlling for political ideology.}
Similar to the setting described above, Task 2 controls for political ideology. 
In particular, we only use debates where the two debaters are from different political ideologies 
({\sc conservative} vs.\ {\sc liberal}). In contrast to Task 1, we consider all voters that 
self-identify with one of the two debater ideologies (regardless of whether the voter's
stance changed post-debate vs.\ pre-debate).  
This time, we predict whether the voter gives more total points to the {\sc pro} side or the {\sc con} side
argument. Thus, Task 2 uses \textit{Criterion 1} to determine the winner of the debate 
from the point of view of the voter.  Our hypothesis is that the voter will assign
more points to the debater that has the same political ideology as the voter.

For this task too, we perform the study for two cases --- debates from the \textit{Politics} 
category only and debates from all categories. And we expect to see stronger effects of prior
beliefs for debates on \textit{Politics}.

\subsection{Features}
The features we use in our model are shown in Table~\ref{table:features}.  They can be
divided into two groups --- features that describe the prior beliefs of the users
and linguistic features of the arguments themselves.

\subsubsection*{User features}
We use the cosine similarities between the voter and each of the debaters' \textit{big issue} vectors. 
These features give a good approximation of the overall similarity of two user's opinions.
Second, we use indicator features to encode whether the religious and political beliefs of the 
voter match those of each of the debaters. 

\subsubsection*{Linguistic features}
We extract linguistic features separately for both the {\sc pro} and {\sc con} side of the debate (combining all the utterances of {\sc pro}  across different turns and doing the same for {\sc con}). Table \ref{table:features} contains a list of these features. It includes features that carry information 
about the style of the language (e.g., usage of modal verbs, length, punctuation), represent different
semantic aspects of the argument (e.g., showing evidence, connotation \citep{feng2011classifying},
subjectivity \citep{wilson2005recognizing}, sentiment, swear word features) as well as
features that convey different argumentation styles (argument lexicon features \citep{somasundaran2010recognizing}. Argument lexicon features 
include the counts for the phrases that match with the regular expressions of argumentation styles such as assessment, authority, conditioning, contrasting, emphasizing, generalizing, empathy, inconsistency, necessity, possibility, priority, rhetorical questions, desire, and difficulty. 
We then concatenate these features to get a single feature representation for the entire debate. 
%

\section{Results and Analysis}
\label{results}
For each of the tasks, prediction accuracy is evaluated using 5-fold cross validation. We pick the model parameters for each split with 3-fold cross validation on the training set. We do ablation for each of user-based and linguistic features. We report the results for the feature sets that perform better than the baseline. 


We perform analysis by training logistic regression models using only user-based features, only linguistic features and finally combining user-based and linguistic features for both the tasks. \\ 
\begin{table}[h]
\begin{tabular}{p{5cm}p{1.7cm}}
		  & \textbf{Accuracy}  \\
\hline
\textbf{Baseline} &   \\

{Majority} & $56.10$\% \\ 
 \hline
\textbf{User-based Features} &   \\
      {Matching religious ideology} & \textbf{$65.37$} \% \\ 
    \hline
    \textbf{Linguistic features} &  \\
    {Personal pronouns} & \textbf{$57.00$} \%\\
    {Connotation} & \textbf{$61.26$} \%\\
    {All two features above} & \textbf{$65.37$} \%\\
    
   \hline
   	\textbf{User-based+linguistic features} &   \\
   {{\sc user*}+ Personal pronouns}  &$65.37$\%\\
   {{\sc user*}+ Connotation}  &$66.42$\%\\
  {{\sc user*}+ {\sc language*}}  &$64.37$\%\\
  \hline
\end{tabular}

\caption{Results for Task 1 for debates in category \textit{Religion}. 
{\sc user*} represents the best performing combination of user-based features. {\sc language*} represents the best performing combination of linguistic features.
Since using linguistic features only would give the same prediction for all voters in a debate, the maximum accuracy that can be achieved using language features only is $92.86$\%. }
\label{table:task1_religion}
\end{table}

\newpage
\textbf{Task 1 for debates in category \textit{Religion}.} As shown in Table \ref{table:task1_religion}, the majority baseline (predicting the winner side of the majority of training examples out of {\sc pro} or {\sc con}) gets $56.10$\% accuracy. User features alone perform significantly better than the majority baseline. The most important user-based feature is \textit{matching religious ideology}. This means it is very likely that people change their views in favor of a debater with the same religious ideology. In a linguistic-only features analysis, combination of 
the \textit{personal pronouns} and \textit{connotation} features emerge as most important and also perform significantly better than the majority baseline at $65.37$\% accuracy. When we use both user-based and linguistic  features to predict, the accuracy improves to $66.42$\% with \textit{connotation} features. An interesting observation is that including the user-based features along with the linguistic features changes the set of important linguistic features for persuasion removing the \textit{personal pronouns} from the important linguistic features set. This shows the importance of studying potentially confounding user-level factors.
\begin{table}[h]
\begin{center}
\begin{tabular}{p{5cm}p{1.7cm}}
  & \textbf{Accuracy}  \\
\hline
\textbf{Baseline} &  \\
{Majority} & $57.31$\% \\ 
 \hline
\textbf{User-based Features} &   \\
{Matching religious ideology} & $62.79$ \% \\ 
 \specialcell[b]{{Matching religious ideology}+ \\ Opinion similarity} & \textbf{$62.97$}\%\\
      
    \hline
    \textbf{Linguistic features} &  \\
    {Length \footnote{This linguistic feature is the one achieving the best performance.}} &  $61.01$ \% \\
    \hline
   	\textbf{User-based+linguistic features} &   \\
  {{\sc user*} + Length}  & $64.56$ \%\\
   \specialcell[b]{{\sc user*}+ Length\\+ Exclamation marks} & {$65.74$}\%\\
 \hline
\end{tabular}
\end{center}
\caption{Results for Task 1 for debates in all categories. 
The maximum accuracy that can be achieved using language features only is $95.77$\%.
}
\label{table:task_1_all}
\end{table}

\textbf{Task 1 for debates in all categories.} As shown in Table 
\ref{table:task_1_all}, for the experiments with user-based features only, \textit{matching religious ideology} and \textit{opinion similarity} features are the most important. For this task, \textit{length} is the most predictive linguistic feature and can achieve significant improvement over the baseline ($61.01$\%). When we combine the language features with user-based features, we see that with \textit{exclamation mark} the accuracy improves to ($65.74$\%).

\textbf{Task 2 for debates in category \textit{Politics}.} As shown in Table \ref{table:task_2_politics}, using user-based features only, the \textit{matching political ideology} feature performs the best ($80.40$\%). Linguistic features (refer to Table \ref{table:task_2_politics} for the full list) alone, however, can still obtain significantly better accuracy than the baseline ($59.60$\%). The most important linguistic features include \textit{approval}, \textit{politeness}, \textit{modal verbs}, \textit{punctuation} and \textit{argument lexicon features} such as \textit{rhetorical questions} and \textit{emphasizing}. 
When combining this linguistic feature set with the \textit{matching political ideology} feature, we see that with the accuracy improves to  ($81.81$\%). \textit{Length} feature does not give any improvement when it is combined with the user features. 

\begin{table}[h]
\begin{center}
\begin{tabular}{p{5cm}p{1.7cm}}
                & \textbf{Accuracy}  \\
\hline
\textbf{Baseline} &  \\ 
{Majority} & $50.91$\% \\ 
 \hline
\textbf{User-based Features} &   \\
  {Opinion similarity} & $80.00$ \% \\
    {Matching political ideology} & \textbf{$80.40$} \% \\
    \hline
    \textbf{Linguistic features} &  \\
    {Length} & $57.37$ \% \\
    \textit{linguistic feature set} & \textbf{$59.60$} \%\\
   \hline
   \textbf	{User-based+linguistic features} &   \\
  {{\sc user*}+ \textit{linguistic feature set}}  & \textbf{$81.81$}\%
   \\
   \hline
\end{tabular}
\end{center}
\caption{Results for Task 2 for debates in category \textit{Politics}. 
The maximum accuracy that can be achieved using linguistic features only is $75.35$\%. 
The \textit{linguistic feature set} includes \textit{rhetorical questions, emphasizing, approval, exclamation mark, questions, politeness, referring to opponent, showing evidence, modals, links,} and \textit{numbers} as features.}
\label{table:task_2_politics}
\end{table}

\begin{table}[h]
\begin{center}
\begin{tabular}{p{5cm}p{1.7cm}}
 & \textbf{Accuracy}  \\
\hline
\textbf{Baseline} &   \\
{Majority} & $51.75$\% \\ 
 \hline
\textbf{User-based Features} &  \\
  {Opinion similarity} & 
 $73.96$\% \\
  
    \hline
   
    \textbf{Linguistic features} &  \\
    
    {Length} & $56.88$\% \\
    {Politeness} & $55.00$\% \\
    {Modal verbs} & $52.32$\% \\
    {Tf-idf features}  &  \textbf{$52.89$} \%\\
   \hline
   	\textbf{User-based+linguistic features} &   \\
  {{\sc user*}+ Length}  & $74.53$\%\\
  {{\sc user*}+ Tf-idf}  & $74.13$\%\\
  \specialcell[b]{{\sc user*}+ Length\\ + Tf-idf} & \textbf{$75.20$}\%\\
  
  \hline
\end{tabular}
\end{center}
\caption{Results for Task 2 for debates in all categories.
The maximum accuracy that can be achieved using linguistic features only is $74.53$\%.
}
\label{table:task_2_all}
\end{table}

\textbf{Task 2 for debates in all categories.}
As shown in Table \ref{table:task_2_all}, when we include all categories, we see that the best performing user-based feature is the \textit{opinion similarity} feature ($73.96$\%). When using language features only, \textit{length} feature ($56.88$\%) is the most important. For this setting, the best accuracy is achieved when we combine user features with \textit{length} and \textit{Tf-idf} features. We see that 
the set of language features that improve the performance of user-based features do not include some of that perform significantly better than the baseline when used alone (\textit{modal verbs} and \textit{politeness} features).

\section{Related Work}
\label{rel-work}
Below we provide an overview of related work from the multiple disciplines that study persuasion. 

\textbf{Argumentation mining.} Although most recent work on argumentation has focused on identifying 
the structure of arguments and extracting argument components 
\cite{PersingN15,
palau2009argumentation,biran2011identifying,mochales2011argumentation,feng2011classifying,stab-gurevych:2014:Coling,lippi2015contextindependent,park-cardie:2014:W14-21,nguyen-litman:2015:ARG-MINING,peldszus-stede:2015:EMNLP,P17-1091,rosenthal2015couldn}, more relevant is research on identifying the characteristics of persuasive text, e.g., what distinguishes persuasive from non-persuasive text \cite{tan2016winning,zhang2016conversational,wachsmuth2016using,habernal2016makes,habernal2016argument,fang2016learning,hidey-EtAl:2017:ArgumentMining}. Similar to these, our work aims to understand the characteristics of persuasive text but also considers the effect of people's prior beliefs. 

\textbf{Persuasion.} 
There has been a tremendous amount of research effort in the social sciences (including computational social science) to understand the characteristics of persuasive text   \cite{kelman1961processes,burgoon1975toward,
chaiken1987heuristic,tykocinskl1994message,
doi:10.1177/0741088396013003001,
dillard2002persuasion,
Cialdini.2007,
durik2008effects,
tan2014effect,
Marquart2016}. Most relevant among these is the research of  \newcite{tan2016winning}, \newcite{habernal2016makes} and \newcite{hidey-EtAl:2017:ArgumentMining}.
\newcite{tan2016winning} focused on the effect of user interaction dynamics and language features looking at the ChangeMyView\footnote{\href{https://www.reddit.com/r/changemyview/}{https://www.reddit.com/r/changemyview/}} (an internet forum) 
community on Reddit and found that user interaction patterns as well as linguistic features are connected to the success of persuasion. In contrast, 
\newcite{habernal2016makes} created a crowd-sourced corpus consisting of argument pairs and,
given a pair of arguments, asked annotators which is more convincing. This allowed them to experiment with different features and machine learning techniques for persuasion prediction. Taking motivation from Aristotle's definition for modes of persuasion, \newcite{hidey-EtAl:2017:ArgumentMining} annotated claims and premises extracted from the ChangeMyView community with their semantic types to study if certain semantic types or different combinations of semantic types appear in persuasive but not in non-persuasive essays. In contrast to the above, our work focuses on persuasion in debates than monologues and forum datasets and accounts for the user-based features. 

\textbf{Persuasion in debates.} 
Debates are another resource for studying the different aspects of persuasive arguments. Different from monologues  where the audience is exposed to only one side of the opinions about an issue, debates allow the audience to see both sides of a particular issue via a controlled discussion. 
There has been some work on argumentation and persuasion on online debates. \newcite{sridhar2015joint}, \newcite{somasundaran2010recognizing} and \newcite{hasan2014you},
for example, studied detecting and modeling stance on online debates.
\newcite{zhang2016conversational} found that the side that can adapt to their opponents' discussion points over the course of the debate is more likely to be the winner.
None of these studies investigated the role of prior beliefs in stance detection or persuasion.\\
\hspace*{1em}\textbf{User effects in persuasion.} Persuasion is not independent from the characteristics of the people to be persuaded. 
Research in psychology has shown that people have biases in the ways they interpret the arguments they are exposed to because of their prior beliefs \cite{lord1979biased,vallone1985hostile,doi:10.1177/0741088396013003001}. Understanding the effect of persuasion strategies on people, the biases people have and the effect of prior beliefs of people on their opinion change has been an active area of research interest \cite{correll2004affirmed,hullett2005impact,petty1981personal}.  \newcite{eagly1975attribution},
for instance, found that the attractiveness of the communicator plays an important role in persuasion. Work in this area could be relevant for the future work on modeling shared characteristics between the user and the debaters.
 To the best of our knowledge, \newcite{lukin2017argument} is the most relevant work to ours since they consider features of the audience on persuasion. In particular, they studied the effect of an individual's personality features (open, agreeable, extrovert, neurotic, etc.)
on the type of argument (factual vs.\ emotional) they find more persuasive. Our work differs from this work since we study debates and in our setting the voters can see the debaters' profiles as
well as all the interactions between the two sides of the debate rather than only being exposed to a monologue. Finally, we look at different types of user profile information such as a user's religious and ideological beliefs and their opinions on various topics. 

\section{Conclusion}
\label{conclusions}
In this work we provide a new dataset of debates and a more controlled setting to study the effects of prior belief on persuasion. The dataset we provide and the framework we propose open several avenues for future research. One could explore the effect different aspects of people's background (e.g., gender, education level, ethnicity) on persuasion. Furthermore, it would be interesting to study how people's prior beliefs affect their other activities on the website and the language they use while interacting  with people with the same and different prior beliefs. Finally, one could also try to understand in what aspects and how the language people with different prior beliefs/backgrounds use is different. These different directions would help people better understand characteristics of persuasive arguments and the effects of prior beliefs in language. 
\section{Acknowledgements}
This work was supported in part by NSF grant SES-1741441 and DARPA DEFT Grant FA8750-
13-2-0015.  The views and conclusions contained herein are those of the authors and should not
be interpreted as necessarily representing the official policies or endorsements, either expressed or
implied, of NSF, DARPA or the U.S. Government.
We thank Yoav Artzi, Faisal Ladhak, Amr Sharaf, Tianze Shi, Ashudeep Singh and the anonymous reviewers for their helpful feedback. We also
thank the Cornell NLP group for their insightful comments. 
\bibliography{naaclhlt2018}
\bibliographystyle{acl_natbib}

\appendix

\end{document}